\documentclass{article}

\usepackage{arxiv}

\usepackage[utf8]{inputenc} 
\usepackage[T1]{fontenc}    
\usepackage{lmodern}        
\usepackage{hyperref}       
\usepackage{url}            
\usepackage{booktabs}       
\usepackage{amsfonts}       
\usepackage{nicefrac}       
\usepackage{microtype}      
\usepackage{lipsum}
\usepackage{graphicx}

\title{Stylistic Fingerprints, POS-tags and Inflected Languages: A Case
Study in Polish}
\def\shorttitle{Stylistic Fingerprints, POS-tags and Inflected Languages}

\author{
    Maciej Eder
   \\
    Institute of Polish Language \\
    Polish Academy of Sciences \\
  Kraków, Poland \\
  \texttt{maciej.eder@ijp.pan.pl} \\
   \And
    Rafał L. Górski
   \\
    Institute of Polish Language \\
    Polish Academy of Sciences \\
  Kraków, Poland \\
  \texttt{rafal.gorski@ijp.pan.pl} \\
  }

\newlength{\csllabelwidth}
\newlength{\cslhangindent}
  {}%
  {\par}
\newenvironment{CSLReferences}[2] 
 {
  \setlength{\parindent}{0pt}
  \ifodd #1 \everypar{\setlength{\hangindent}{\cslhangindent}}\ignorespaces\fi
  \ifnum #2 > 0
  \setlength{\parskip}{#2\baselineskip}
  \fi
 }%
 {}
\usepackage{calc} 

\usepackage{longtable}

\begin{document}
\maketitle

\def\tightlist{}

\begin{abstract}
In stylometric investigations, frequencies of the most frequent words
(MFWs) and character \emph{n}-grams outperform other style-markers, even
if their performance varies significantly across languages. In inflected
languages, word endings play a prominent role, and hence different word
forms cannot be recognized using generic text tokenization. Countless
inflected word forms make frequencies sparse, making most statistical
procedures complicated. Presumably, applying one of the NLP techniques,
such as lemmatization and/or parsing, might increase the performance of
classification. The aim of this paper is to examine the usefulness of
grammatical features (as assessed via POS-tag \emph{n}-grams) and
lemmatized forms in recognizing authorial profiles, in order to address
the underlying issue of the degree of freedom of choice within lexis and
grammar. Using a corpus of Polish novels, we performed a series of
supervised authorship attribution benchmarks, in order to compare the
classification accuracy for different types of lexical and syntactic
style-markers. Even if the performance of POS-tags as well as lemmatized
forms was notoriously worse than that of lexical markers, the difference
was not substantial and never exceeded \emph{ca}. 15\%.
\end{abstract}


\hypertarget{introduction}{%
\section{Introduction}\label{introduction}}

In classical approaches to authorship attribution, frequencies of the
most frequent words (MFWs) and other style-markers such as character
\emph{n}-grams are claimed to outperform other types of style-markers
(Koppel et al., 2009; Stamatatos, 2009), even if their performance
varies significantly across different languages (Eder, 2011; Rybicki and
Eder, 2011; Evert et al., 2017). Also, it has been proven that
attribution based on single words and, even more so, on letter
\emph{n}-grams reveal a very high resistance to errors in corpora such
as those derived from imperfect OCR (Eder, 2013). A previous study in
authorship attribution performed on a large corpus of Polish novels
(Rybicki, 2015a) confirmed the usefulness of most frequent words.
Defined, for any text analysis software, as simple strings of letter-
and non-letter characters, all these plain features are easily extracted
from input texts. One should not underestimate the implications of such
an efficient combination of simplicity and performance. Namely, a
stylometric test -- be it authorship attribution or a distant-reading
analysis of literature using quantitative methods -- can be applied to
any web-scraped plain text file with a high probability of achieving
acceptable results.

Very attractive as they are, these shallow text features have also their
limitations. Firstly, there is little theory that would explain the
phenomenon of the visibility of authorial stylometric signal of the very
frequent features -- apart from the general and intuitive assumptions
that authors might be possessed of their own ``stylistic fingerprint''
(Kenny, 1982: 12) or that the very frequent words, for instance, might
define authorial style by establishing the context for the less frequent
yet more ``meaningful'' words (McKenna et al., 1999). Certainly, there
exist studies that aim to provide a convincing theoretical background
for stylometry (Kestemont, 2014), nevertheless one can say that we are
still at the beginning of the journey. This lack of theory might be the
reason why many scholars look askance at frequency-based quantitative
analyses and, consequently, that there is little dialogue between
quantitative and qualitative approaches to textual analysis.

Secondly, the above-cited findings (Eder, 2011; Rybicki and Eder, 2011;
Evert et al., 2017) cast doubt whether the appropriateness of a
quantitative frequency-based method developed for one language easily
translates into similar success in another, as has been suggested in
earlier studies (Juola, 2009). In fact, the high discrepancy in
authorial attribution success observed in the 2013 experiment has been
suspected by the researchers to stem from the differences in the
inflection of the languages compared. The observation that highly
inflected Polish fared worst among less inflected languages such as
English or German, will be quite relevant in the context of the present
study.

To explore further the hypothesis of inflection's role in attribution:
it is obvious that, in inflected languages, different forms of the same
word cannot be recognized using generic text tokenization (e.g., via
regular expressions). This is a possible source of error, since, in
languages such as Polish, word endings play a prominent role; as a
result, much of the grammatical information that is easily available in,
say, English function words, remains ``hidden'', or ``dissolved'', in
inflected nouns or verbs, and has no way of making it to the top ranks
in frequency lists: these countless inflected word forms make word
frequencies sparse, and this complicates most statistical procedures.

Meanwhile, morphologically rich languages with relatively free
word-order, such as Polish, are significantly different from the
grammatical point of view, and it should not come as a surprise that
they make the task substantially different. With its 7 cases multiplied
by 2 numbers, singular and plural (and, to make things even more
complicated, vestiges of the dual coexisting with the plural, as in
\emph{oczyma}~:~\emph{oczami} `eyes' instrumental), a Polish noun might
have up to 14 different inflected forms. As if this were not enough,
nouns with two alternative endings for some cases are not infrequent
(e.g., \emph{reżyserzy}~:~\emph{reżyserowie} `film directors' nominative
plural). This figure is multiplied when adjectives are concerned, since
they inflect by case, by number, and by gender: there are three genders
in the singular and two in the plural (+human masculine vs.~--human
masculine). And while regular homonymy within the inflection paradigm of
the Polish languages keeps those inflected forms considerably lower than
the above-presented worst-case-scenario, this comes at the cost of a
greater degree of ambiguity. The same general rule holds for verbs,
pronouns and numerals.

This brings us back to the question of authorial attribution, this time
in the distinct context of rich inflection. Presumably, the problem of
the morphological abundance can be overcome -- at least to some extent
-- by lemmatization, or transforming the original sequence of words into
their base forms, as in the following example: \emph{w jednym z
pomniejszych miast perskich mieszkali dwaj bracia} (original words),
\emph{w jeden z pomniejszy miasto perskie mieszkać dwa brat} (lemmatized
words). From a theoretical point of view, the difference between a
sequence of original forms and lemmatized words is not as big as it
might seem. After all, any stylometric inference based on word
frequencies means in fact reducing a very complex phenomenon -- the
natural language -- into its simple representation, while filtering out
a vast amount of original information. Lemmatization is no different in
this respect, except that it reduces the language even more, by cutting
off grammatical information held by the original word forms.

Being an obvious remedy for data sparseness, lemmatization should
increase the visibility of the authorial signal. However, an opposite
hypothesis is also plausible, namely one can assume that the
(grammatically richer) original word forms preserve a cleaner authorial
signature than the grammar-less lemmas. Finally, a hypothesis that the
signal is hidden \emph{between} the original forms and the lemmas --
i.e.~in the grammatical structure itself -- cannot be ruled out. From a
linguistic point of view, this third scenario is rooted in fundamental
questions of the authorial freedom of choice vs.~constrains of the
language.

In principle, grammar will always constrain the authorial freedom of
choice to a significantly greater degree than it constrains the (usually
very individual) lexical repertoire. If an author wishes to describe a
given entity with an adjective, there exist numerous words to choose
from: e.g.~the entity's size may be \emph{big}, \emph{large},
\emph{great}, \emph{considerable} etc. However, if we take into account
grammatical categories, the entity will inevitably be represented by a
sequence {[}Adjective{]} + {[}Noun{]}. Moreover, despite some
limitations in combining words (such as the impossible Chomskyan
\emph{green dreams}), these limitations are much more rigid on the
syntactic level than on the lexical level: once a transitive verb is
introduced, it has to be followed by an object. Additionally, the case
of the object cannot be freely chosen -- it is assigned by the verb.
Therefore, we can easily formulate a pre-empirical assumption that
authors enjoy much larger freedom of choice on the level of lexis
compared to syntax. Certainly, novelists usually try to be creative and
do not adhere to most typical collocations\footnote{Our corpus contains
  literary sources only. An interesting question -- far beyond the scope
  of this study -- is the extent to which the fact that a writer seeks
  originality makes the fingerprint clearer compared to non-fiction
  literature.}, but even a highly experimental artistic novel cannot
ignore language constrains.

It is quite clear, then, that grammar should not be excluded from the
experimental setup of the present study. Yet, the problem of extracting
the grammatical structure from texts (referred to as parsing) is far
more complex than lemmatization. It is true that, despite new
developments in this area, automatic parsing is still somewhat
unreliable and obtaining a tailor-made tree bank is beyond our
capabilities; however, straightforward insight into grammar can be
obtained using Part-of-Speech (POS) tags combined into \emph{n}-grams
(Wiersma et al., 2011). Attempts to solve this problem have already
yielded promising results (Baayen et al., 1996; Hirst and Feiguina,
2007) -- yet, once again, mostly in English.

The downside of such an approach is that the POS \emph{n}-grams can
provide us with a rather rough model of syntax or, in the words of
Wiersma et al. (2011), ``a good aggregate representation of syntax''.
However, since these features were compared in the context of repetitive
authorial decisions -- conscious or unconscious -- that make texts by
the same author more similar to each other than to texts by other
authors, there was some hope that such an experiment might provide an
insight to the various degrees of linguistic choice at the lexical
and/or syntactic level.

Because of the complexity of individual word forms' grammatical
information, morphologically rich languages are usually annotated with
so-called positional tags, i.e.~sequences of codes for all the values of
grammatical categories which pertain to a word, where only one segment
of a tag stands for the part of speech itself. To illustrate, while the
English word \emph{impossible} is tagged \texttt{AJ0} (Adjective,
general or positive), the Polish \emph{niemożliwemu}, the Dative
Singular of the same adjective \emph{impossible}, must be described by a
fairly verbose tag: \texttt{adj:sg:dat:m1:pos}, where ``adj'' stands for
Adjective, ``sg'' for singular, ``dat'' for Dative, ``m1'' for
Masculine-Virile, ``pos'' for Positive Grade. Consequently, this complex
tag is a bundle of inflectional features of the word; its code for case,
number, and gender \texttt{sg:dat:m1} can also form a part of a
substantive or participle, whereas the first segment of the sequence,
``adj'', is the only part of the tag that is directly comparable to its
English counterpart.

Arguably, a Polish unlemmatized text has a much lower type/token ratio
than a lemmatized one. Equally obviously, a comparable English text (for
instance, an English translation of a Polish text) produces a lower TTR.
Finally, the difference of TTR in a lemmatized and unlemmatized English
text is much less prominent. In the context of automatic POS tagging,
the difference accounts for a substantial increase in the difficulty of
this task as the rich morphology in Polish requires a vast number of
tag-types. The tagset of the National Corpus of Polish (Przepiórkowski
et al., 2012) amounts to over 1,000 tags, a full degree of magnitude
more than the mere 140 tags in the CLAWS-8 tagset for English. This
means, among other things, that a Polish POS-tagged text would produce
much lower frequencies for every POS type. And if this were not enough,
the relatively free word order in Polish makes one expect a higher
number of different POS-tag combinations (\emph{n}-grams), since a
sequence of two or more parts of speech can occur in different order. It
is true that several restrictions on Polish word order might slightly
attenuate this phenomenon, e.g.~the preposition can never be placed in
postposition, and the negation of the verb must immediately precede the
latter, nevertheless the increase in the number possible POS-tag
\emph{n}-grams is still remarkable.

\hypertarget{hypothesis}{%
\section{Hypothesis}\label{hypothesis}}

With all the above remarks taken in the consideration, we can now
formulate the research questions to be addressed in this study: firstly,
we aim to empirically examine the amount of authorial signal that
resides in grammar as assessed through POS-tags; secondly, we aim at
comparing the performance of original word forms against lemmatized
forms (a scenario in which \emph{some} of the grammatical information is
stripped out). Additionally, we aim to test the extent to which
particular segments of positional tags (analyzed separately and combined
into \emph{n}-grams) might be useful in this respect. Therefore, apart
from the entire tags, their segments have also been assessed, namely
\emph{n}-grams of single categories, as well as combinations of two tag
segments. In this approach the Polish word sequence \emph{jedną czerwoną
ranę} (\emph{one red wound} in the Accusative form) was analyzed as word
forms, as lemmas (e.g.~\emph{jeden czerwony rana}), and as different
chains of POS-tag parts:

\begin{enumerate}
\def\labelenumi{\arabic{enumi}.}
\tightlist
\item
  entire tags, e.g.~{[}adj:sg:acc:f:pos{]} + {[}adj:sg:inst:f:pos{]} +
  {[}subst:sg:acc:f{]};
\item
  POS tags in the strict sense, or the first segments only,
  e.g.~{[}adj{]} + {[}adj{]} + {[}subst{]};
\item
  tags cut off after their second segment, e.g.~{[}adj:sg{]} +
  {[}adj:sg{]} + {[}subst:sg{]}.
\end{enumerate}

The above word forms, lemmas and different variants of grammatical tags
were further combined into \emph{n}-grams (ranging from 1-grams to
3-grams), resulting in 15 distinct types of style-markers assessed
individually in controlled authorship attribution tests.

Our working hypothesis is that the grammatical layer will exhibit some
traces of authorial signal, yet they will not overshadow the primary
signal produced by the lexical layer. As for the lemmatized
vs.~unlemmatized words as efficient style-markers, we hypothesize that
an input text partially stripped out of grammar, i.e.~lemmatized, will
exhibit a slightly stronger authorial voice compared to original word
forms.

\hypertarget{data-and-method}{%
\section{Data and method}\label{data-and-method}}

In order to corroborate the above hypotheses, we compiled a tailored
corpus of 189 novels in Polish. It is true that restricting the choice
to exclusively one genre (literary novels) will not allow us to
generalize the results to the Polish language in its entirety, However,
we wanted to control for genre in our experiments, as it is usually a
crucial factor in authorship attribution. Similarly, we choose novels
because of their naturally large size, which will prevent the authorial
signal from being blurred by the short sample effect.

The corpus consists of Polish novels from the 20\textsuperscript{th}
century, all of them drawn from the National Corpus of Polish. They were
processed using the Pantera tagger (Acedański, 2010) fully
automatically. No stop lists were used, punctuation marks were treated
on a par with words, certainly the same holds for POS-tags. The full
dataset consisted of 189 Polish novels written by 46 authors; each
author was represented by 3 to 6 texts (4.1 on average). The
chronological range was maintained to be possibly narrow, in order to
minimize the potential impact of diachronic linguistic change; it has
been reported that chronology is a strong signal in most-frequent-word
based stylometry (Burrows, 1996; Rybicki, 2015b). Smaller subsets of the
main corpus were also analyzed in two additional cross-check
experiments, one involving 99 novels by 33 authors, and the other 30
novels by 10 authors (in both setups, the even number of 3 books per
author was secured). Due to copyright restrictions, the novels used in
this study could not be made publicly available. However, we post all
frequency tables used in this study, as well as the full set of the
results, followed by the code needed to replicate the tests, on our
GitHub repository:
\url{https://github.com/computationalstylistics/PL_lemmatization_in_attribution}.

In all, 5 different variants of features were tested for attribution
success: (1) unlemmatized words (original word forms); (2) lemmatized
words; (3) full tags; (4) POS-tags in the strict sense, i.e.~the labels
of the Part of Speech alone; (5) two initial tag parts. All these were
analyzed in \emph{n}-grams, at \emph{n} from 1 to 3; which resulted in
15 independent classification experiments. The analyses were performed
for 35 features, and then for 100, 150, 200 and onward up to 2,000 most
frequent items, by increments of 50. Finally, four supervised
machine-learning classifiers were compared: Burrows's Delta, Cosine
Delta, Support Vector Machines (SVM), and Nearest Shrunken Centroids
(NSC). The entire experimental setup was repeated for the three variants
of the corpus, comprising of 189, 99 and 30 novels, respectively.

The choice of the four classification methods was based on their
time-proven applicability to solving authorship attribution tasks.
Delta, a simple distance-based method introduced by Burrows (2002),
enjoys a reasonable share of attention in stylometry due to its
simplicity and efficiency. Next comes its variant known as the Cosine
Delta, which has been proven to outperform most of distance-based
classifiers (Evert et al., 2017). The Nearest Shrunken Centroids,
another distance-based learner, has also been successfully applied to
text classification (Jockers and Witten, 2010). Support Vector Machines
is a widely-known multidimensional classifier, commonly believed to be
one of the best machine-learning techniques for data analysis. It has
been shown that the performance of this method is very high indeed
(Koppel et al., 2009; Jockers and Witten, 2010). In our approach we use
a simple SVM setup: linear kernel (rather than polynomial) with the cost
parameter set to 1 (rather than optimized in cross-validation). While
parameter tuning usualy improves the performance of SVM, we aimed at
keeping the experimental conditions identical for all analyzed
scenarios.

One has to emphasize, however, that the classification setup we deal
with here is substantially different from typical attribution problems,
since it involves multiple classes, instead of the usual two or three.
Such a situation is referred to as the ``needle in a haystack''
attribution scenario (Koppel et al., 2009), i.e.~a type of attribution
in which the real author is hidden among a very high number of false
candidates. An obvious question arises whether a multi-class setup --
significantly more demanding than a standard attribution experiment --
is a good choice to assess the performance of different style-markers in
a given corpus. An answer to this question is twofold. Firstly, it must
be remembered that, since there are not so many prolific authors, the
number of available texts is also limited; moreover, the access to
electronic versions of those texts is also restricted. Our corpus is no
exception -- the main criterion of including particular texts was their
availability. Secondly and more importantly, a corpus of diverse
authors, authors' genders, genres, topics, audience targets etc.
eliminates possible biases which we can easily overlook. Above all,
however, we should emphasize that we did not aim to improve the overall
accuracy in absolute terms. Rather, we aimed at comparing the efficiency
of several style-markers under identical conditions of the experiment.

The analyses were done using a custom script for R, based on the
\texttt{crossv()} function of the \texttt{stylo} package (Eder et al.,
2016). Particular combinations of style-markers, \emph{n}-grams, and
classifiers, were assessed independently. The scores for subsequent
ranges of the most frequent items were recorded in a leave-one-out
cross-validation scenario. In such a case, all the texts but one were
put into the training set, and the remaining single sample was
classified against the training set. The same procedure was performed
iteratively over the corpus, in each iteration a subsequent text (one at
a time) being excluded for classification. The resulting row of
predicted classes was then compared against the expected classes, and
the number of correct ``guesses'' was recorded as the model's general
accuracy.

Being conceptually very simple and compact, however, accuracy is
considered to overestimate the actual classification performance. For
this reason, a routinely applied toolbox of measures not only includes
accuracy, but also recall, precision, and particularly the F1 score. The
reason why these somewhat less intuitive measures are often neglected in
stylometric studies, is that they are not designed for assessing
multi-class scenarios. Since in our experiment 46 authorial classes were
involved, we relied on \emph{macro-averaged} versions of precision,
recall and the F1 score (Sokolova and Lapalme, 2009). Keeping in mind
that the F1 score in a way combines the information provided by both
recall and precision, this will be our primary diagnostic measure
hereafter.

\hypertarget{results}{%
\section{Results}\label{results}}

The high number of particular attribution tests for different
classification methods, features, \emph{n}-grams, and datasets, calls
for a structured way of presenting the results. For this reason, we will
start with a manual inspection of a somewhat random subset of outcomes.
We will then summarize the differences between the three datasets, the
four classifiers, and finally, we will discuss the performance of
particular style-markers: original words, lemmas and POS-tags.

A small subset of the results is presented in Table 1. Here we report
the performance for the full corpus of 189 novels, original word forms
(MFWs), \emph{n}-gram size set to 1 (i.e.~single words), the Cosine
Delta classifier, and 8 different vectors of the most frequent
features\footnote{The full set of tables for particular datasets,
  classifiers, feature types, and their \emph{n}-grams, can be found in
  our GitHub repository.}. At a glance, one can identify a sweet spot of
performance at 200 MFWs, but a broader picture shows that similar local
areas of better (or worse) performance are not infrequent. In fact, the
classifier reaches its plateau of optimal performance at around 700
MFWs, to slightly decrease for the vectors of more than 1,200 MFWs.

Table 1. 189 novels, Cosine Delta, most frequent words.

\begin{longtable}[]{@{}lrrrr@{}}
\toprule
features & accuracy & precision & recall & F1 score\tabularnewline
\midrule
\endhead
35 & 0.687 & 0.640 & 0.662 & 0.637\tabularnewline
100 & 0.825 & 0.821 & 0.816 & 0.805\tabularnewline
150 & 0.867 & 0.878 & 0.857 & 0.855\tabularnewline
200 & 0.888 & 0.911 & 0.890 & 0.885\tabularnewline
250 & 0.857 & 0.883 & 0.860 & 0.848\tabularnewline
300 & 0.862 & 0.884 & 0.866 & 0.858\tabularnewline
350 & 0.873 & 0.889 & 0.879 & 0.872\tabularnewline
400 & 0.878 & 0.899 & 0.885 & 0.882\tabularnewline
\ldots{} & \ldots{} & \ldots{} & \ldots{} & \ldots{}\tabularnewline
\bottomrule
\end{longtable}

Due to the obvious limitations of presenting the results in a tabular
format, below we present the outcomes in a form of compact plots, so
that reasonable amounts of information can be shown concurrently. To
further increase the clarity of the plots, we will report the F1 scores
only, while delegating all the remaining measures to the GitHub
repository.

Conveniently, the comparison starts with an overview of the three
datasets we used in our study, i.e.~the corpora of 189, 99, and 30
novels, respectively. As it turns out, the general outcome of the
experiment depend, in good accord with intuition, on the size of the
corpus. The best average scores were obtained for the 30-novel subset;
the subcorpus of 99 novels fared somewhat less well; the performance of
the entire set of 189 novels, however, turned out to be very similar to
that of 99 novels (Fig. 1). As a whole, the results were poorer than
expected, even taking into account the fact that the big number of 46
authorial classes -- the needle-in-a-haystack scenario -- was
responsible for this effect. For the subset of 99 texts by 33 authors,
the highest F1 score achieved was as high as 0.91 for the most effective
set of input parameters. For the entire set of 189 novels, the highest
observed score was 0.92. For 30 novels by 10 authors, the score of 1 was
reached for some style-markers combined with Cosine Delta and, to a
lesser extent, with NSC.

\begin{figure}
\centering
\includegraphics[width=1\textwidth]{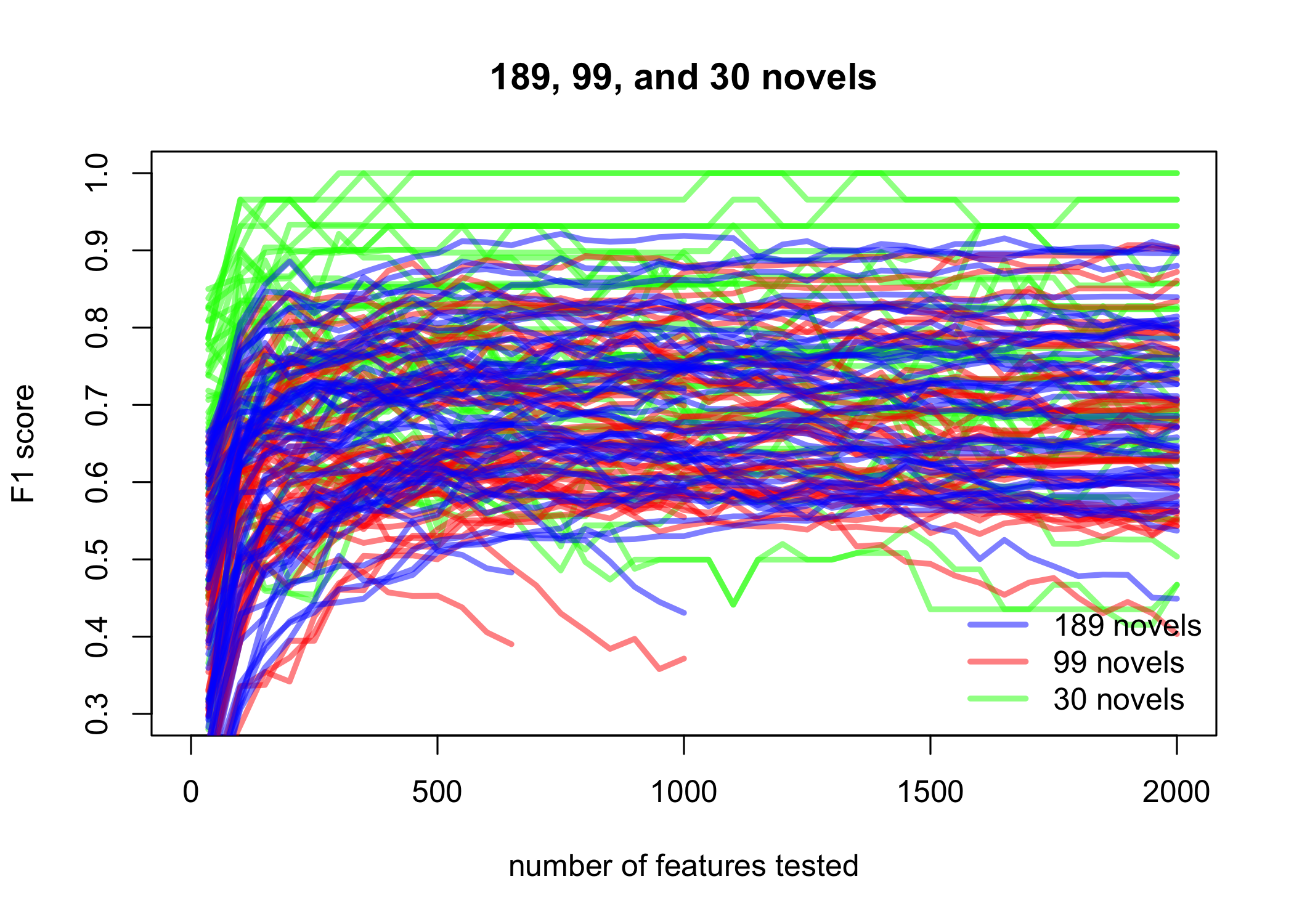}
\caption{Overall performance (F1 scores) for three datasets of
189, 99, 30 novels. Particular curves represent all the style-marker
types and all the classifiers.}
\end{figure}

Despite being compact, the resulting plot (Fig. 1) is rather difficult
to read. For this reason, the information to be plotted will be further
reduced in the next figures. The undeniable collinearity between the
thee corpora of 189, 99, and 30 novels -- despite a few notable
exceptions that will be discussed below -- allow us to focus exclusively
on a single dataset. Therefore, in the following sections we will show
the behavior of the 189-novel corpus alone, delegating the all the
remaining results to the GitHub repository.

\begin{figure}
\centering
\includegraphics[width=1\textwidth]{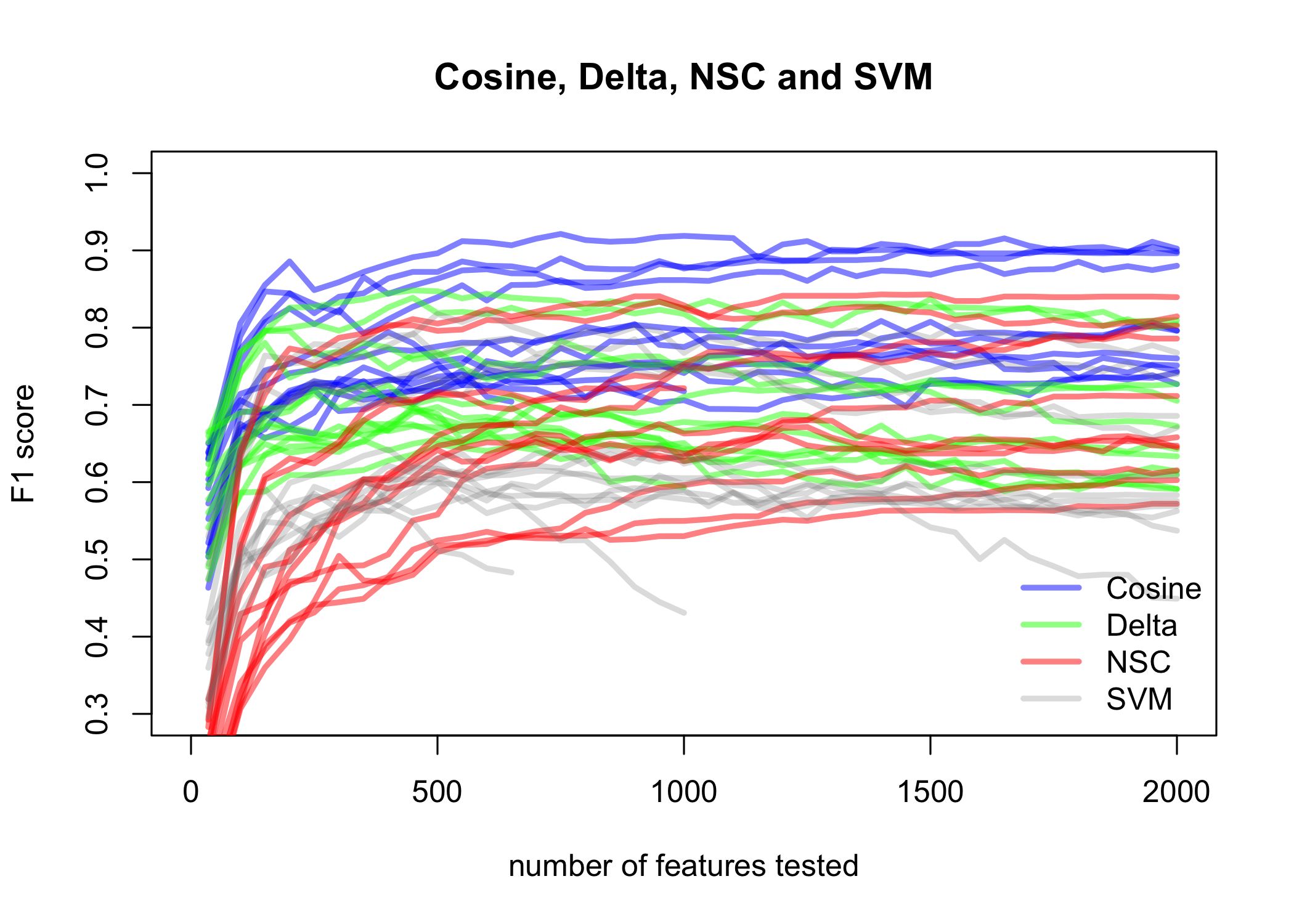}
\caption{Overall performance (F1 scores) for the dataset of 189
novels and four different classifiers: Classic Delta, Cosine Delta, SVM
and NSC.}
\end{figure}

The next comparison was that between our four classifiers. As shown in
Fig. 2, the curves representing performance for each of the classifiers
tend to differ significantly. The top (blue) lines are those for Cosine
Delta, outperforming all the other techniques, as evidenced in recent
scholarship as well (Evert et al., 2017). Next go the performance curves
for Classic Delta that, up to \emph{ca}. 150-word vectors, run together
with those for SVM; but then more and more Classic Delta curves come to
the fore while those for SVM (gray) show a decrease in performance. NSC
exhibits its full potential when long vectors of features are concerned,
which stands in contrast with the behavior of SVM -- while NSC seems to
struggle when the feature space is limited, SVM feels overwhelmed by the
abundance of features. Delta's overall good performance (in both Classic
and Cosine variants) can be partially explained by the fact that in
multi-class setups, distance-based methods usually outperform SVM
(Luyckx and Daelemans, 2011).

Next comes the comparison of particular style-markers' types. The main
research question here is whether lemmatization improves the accuracy of
classification. In Fig. 3, for Cosine Delta, the classical frequent word
approach (MFWs) is highlighted, while all the other curves are kept in
the background. Most frequent word 1-grams (on top) are followed by
2-grams, and then 3-grams (at the bottom). As can be observed, this
simple and time-proven type of features turns out to be the clear winner
of the experiment, at least when Cosine Delta and 189-novel dataset is
concerned. At the same time, however, the same features combined into
3-grams turn out to be unsatisfactory as style-markers, reaching the F1
rate of \emph{ca}. 0.77. There is an explanation of this phenomenon:
being highly inflected, Polish has also a free word-order, which
exponentially increases the number of available word 3-grams (let alone
wider \emph{n}-grams) and leads to substantial data sparseness.

\begin{figure}
\centering
\includegraphics[width=1\textwidth]{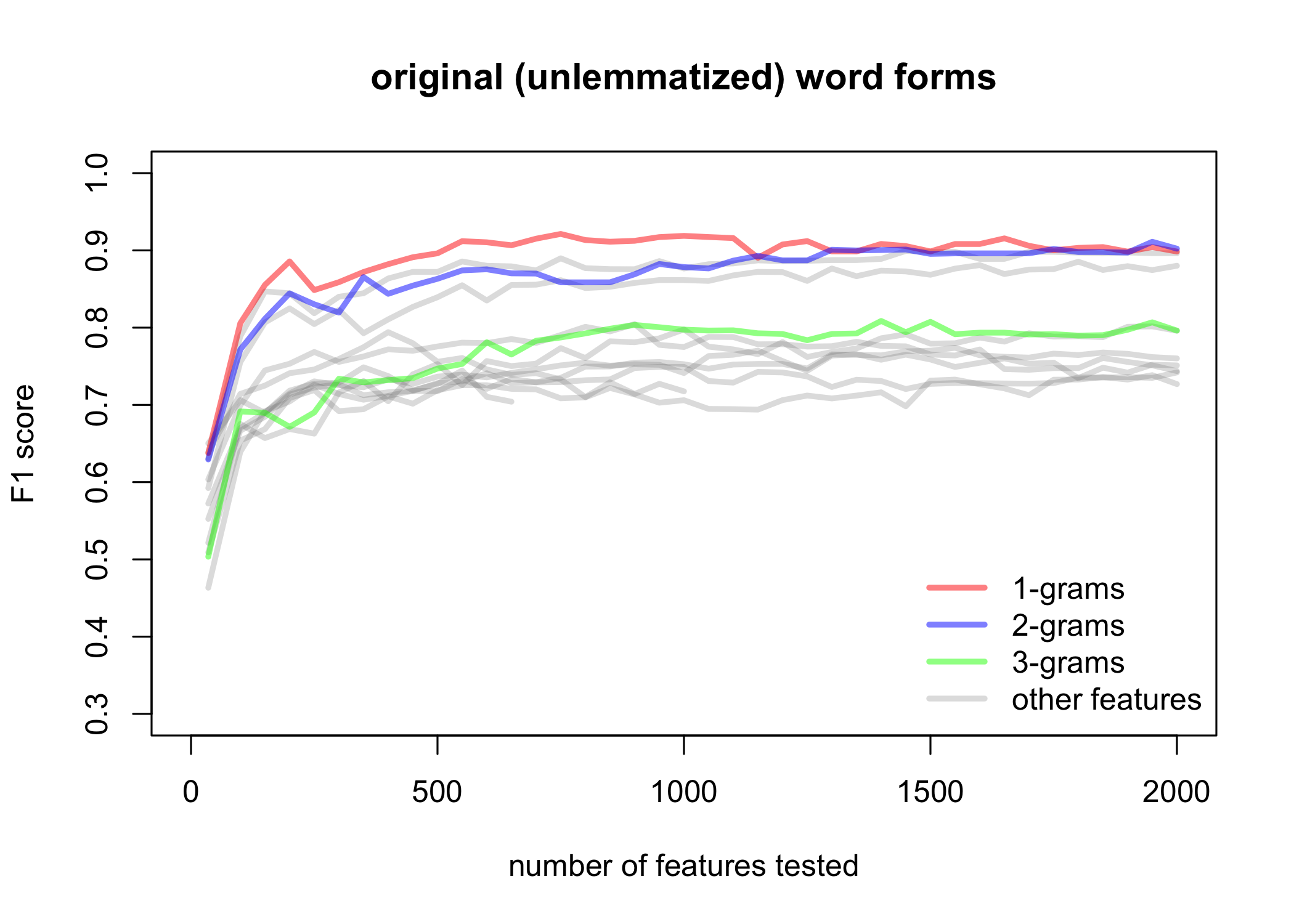}
\caption{Performance of original word forms (unlemmatized words,
or MFWs) in the corpus of 189 novels, assessed by Cosine Delta.}
\end{figure}

Being the best performer, however, frequent word 1-grams are followed
very closely by their immediate competitor, i.e.~frequent lemmatized
word 1-grams (Fig. 4). The general picture of the lemmatized words is
very similar to that of the unlemmatized ones, in terms of both the
dispersion between particular \emph{n}-grams, and the sequence of the
curves: 1-grams are on top, then go 2-grams, while 3-grams are below any
acceptance level. Another noteworthy observation is the fact that both
lemmatized and unlemmatized 1-grams (and 2-grams, to a lesser extent),
rise well above the 0.8 line, which serves as the (mostly unattainable)
ceiling for other style-markers.

\begin{figure}
\centering
\includegraphics[width=1\textwidth]{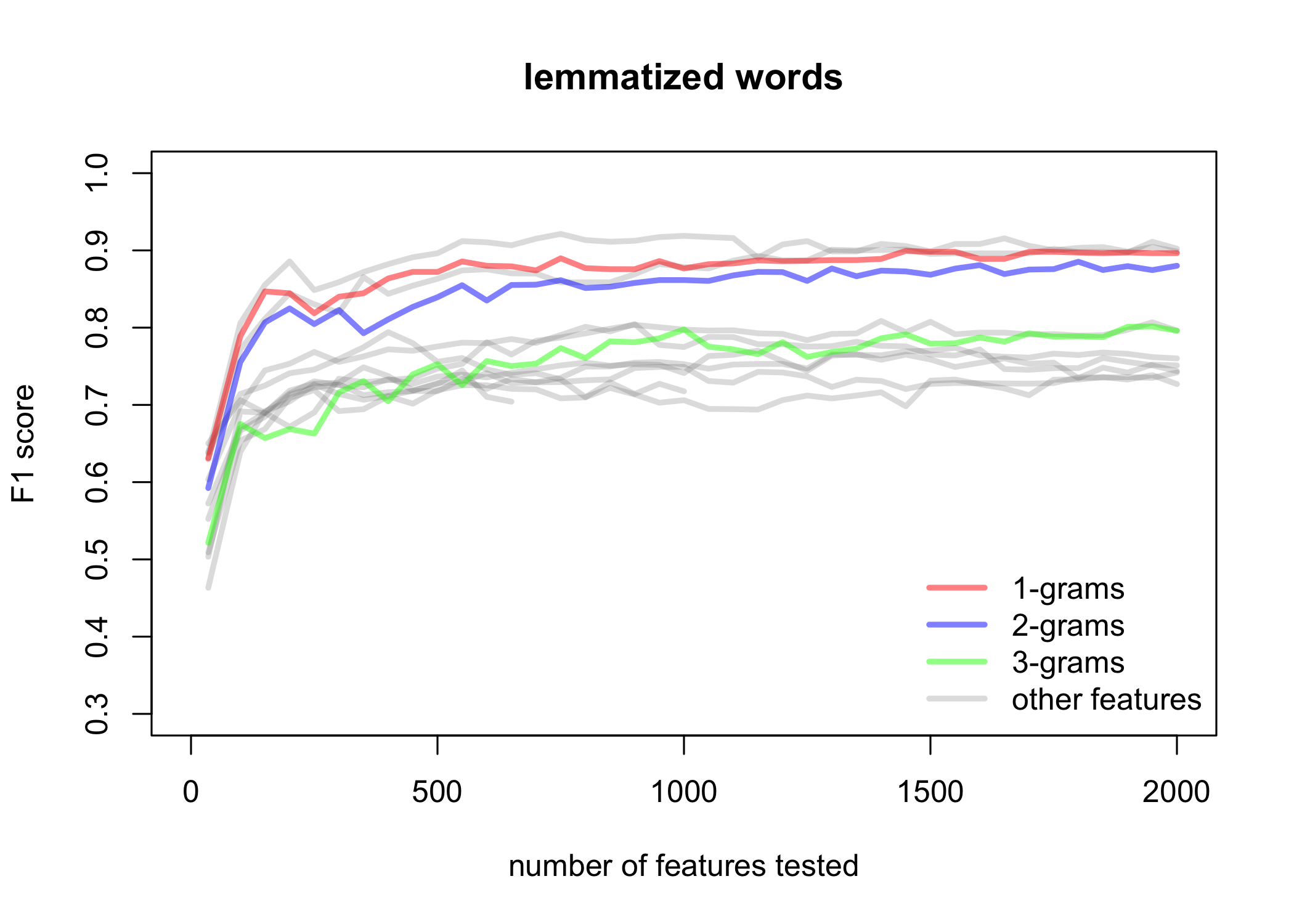}
\caption{Performance of lemmatized words in the corpus of 189
novels, assessed by Cosine Delta.}
\end{figure}

However, the rather small divergence between the lemmatized and
unlemmatized words calls for further exploration. Even if manual
inspection of the respective curves (Fig. 3--4) shows that one of the
style-markers outperforms the other, rigorous statistical testing might
suggest otherwise. A standard way to compare two independent variables,
is to scrutinize them using the t-test. In our case, however, the
variables in question don't meet the formal requirements for t-testing,
since neither of them follows the normal distribution, and their
variances differ significantly. In such a case, Wilcoxon test should be
used instead. According to Wilcoxon test, the difference between
lemmatized and unlemmatized words (for Cosine Delta and the dataset of
189 novels), is indeed significant with a marginally low \emph{p}-value
\textless0.00001. The results of a systematic series of tests for each
combination of the classification method and the dataset are provided in
Table 2. In most cases, the unlemmatized words (MFWs) outperform the
lemmatized words to a significant degree, the exception being the
dataset of 30 novels. Here, a clear winner of the competition cannot be
pointed out, at least for Cosine Delta and NSC.

Table 2. Difference between the F1 scores for grammatical word 1-grams
(i.e.~MFWs) and lemmatized word 1-grams (i.e.~lemmas), assessed by means
of Wilcoxon tests for each combination of classifiers and datasets. The
numbers represent the \emph{p}-values obtained in each individual test.
The asterisks indicate conventional levels of significance.


\begin{tabular}[]{@{}lllll@{}}
\hline
corpus & Delta & Cosine & SVM & NSC\\
\hline
189 novels & 0.000*** & 0.000*** & 0.000*** & 0.000***\\
99 novels & 0.000*** & 0.000*** & 0.004** & 0.002**\\
30 novels & 0.051 & 0.248 & 0.003** & 0.698\\
\hline
\end{tabular}

Finally, the behavior of syntactic style-markers -- as assessed via
POS-tag \emph{n}-grams in their various flavors -- should be commented
on. First and foremost, they turned out to be substantially different in
comparison to lexical markers. As shown in Fig. 5, the overall
performance of full POS-tags is worse than both lemmatized and
unlemmatized words. Also, the spread of the POS-tag curves for different
\emph{n}-grams is smaller (the curves are rather flat) than that of
words, which suggests that the POS-tags are more robust (but also more
resistant to hyperparameter fine-tuning) than lexical markers. Last but
definitely not least, worth noticing is the performance of particular
\emph{n}-grams as a function of the number of features tested. Unlike
the lexical markers, full POS-tag 1-grams don't outperform longer
\emph{n}-grams. It is true that 1-grams initially win, but they are
immediately overtaken by 2-grams, and then even by 3-grams. More
interestingly, the 1-grams reveal a further (and constant) decrease of
performance, as if longer feature vectors contained more and more
stylometric noise.

\begin{figure}
\centering
\includegraphics[width=1\textwidth]{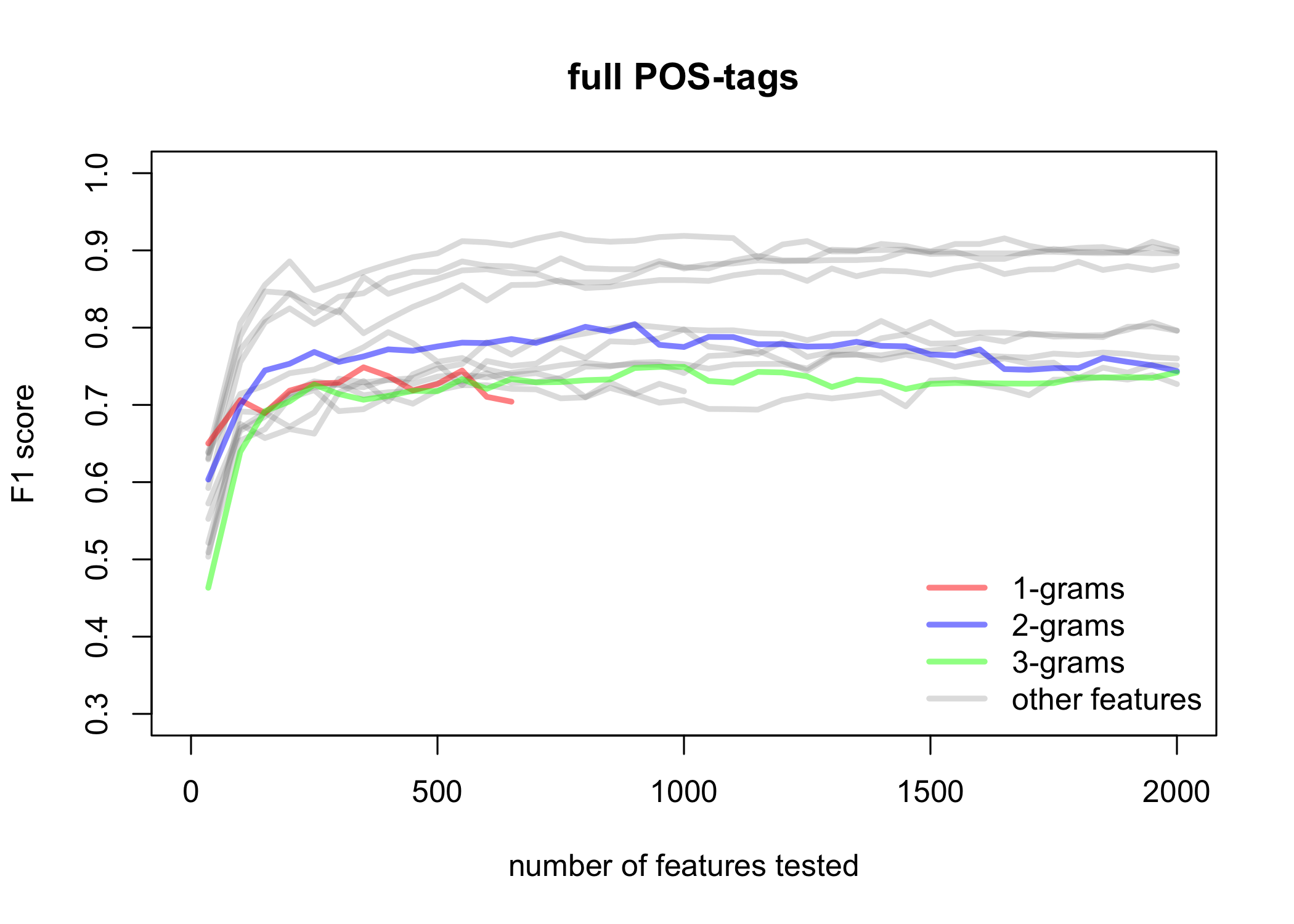}
\caption{Performance of full POS-tags in the corpus of 189
novels, assessed by Cosine Delta.}
\end{figure}

The above picture of syntax-based attribution is corroborated by the
other variants of POS markers, particularly POS-tags in the strict sense
(or, the first tag-parts alone), as shown in Fig. 6. Here, 2-grams
proved optimal, but they reveal a constant decrease of performance for
longer vectors of features, until they are overtaken by 3-grams (the
success rate of 1-grams could be assessed only for the vector of 35
features, reaching the F1 score of 0.643, whereas the number of
available 2-grams was exhausted at the 1,000 features mark). The
behavior of POS-tags reduced to their 1\textsuperscript{st} and
2\textsuperscript{nd} segment (Fig. 7) confirms the general picture of
syntactic features, except that the 3-grams turned out to be the least
successful style-markers examined in this study. Worth mentioning is the
fact that even the the worst choice of features would still lead to the
impressive attributive score of \emph{ca}. 0.75.

\begin{figure}
\centering
\includegraphics[width=1\textwidth]{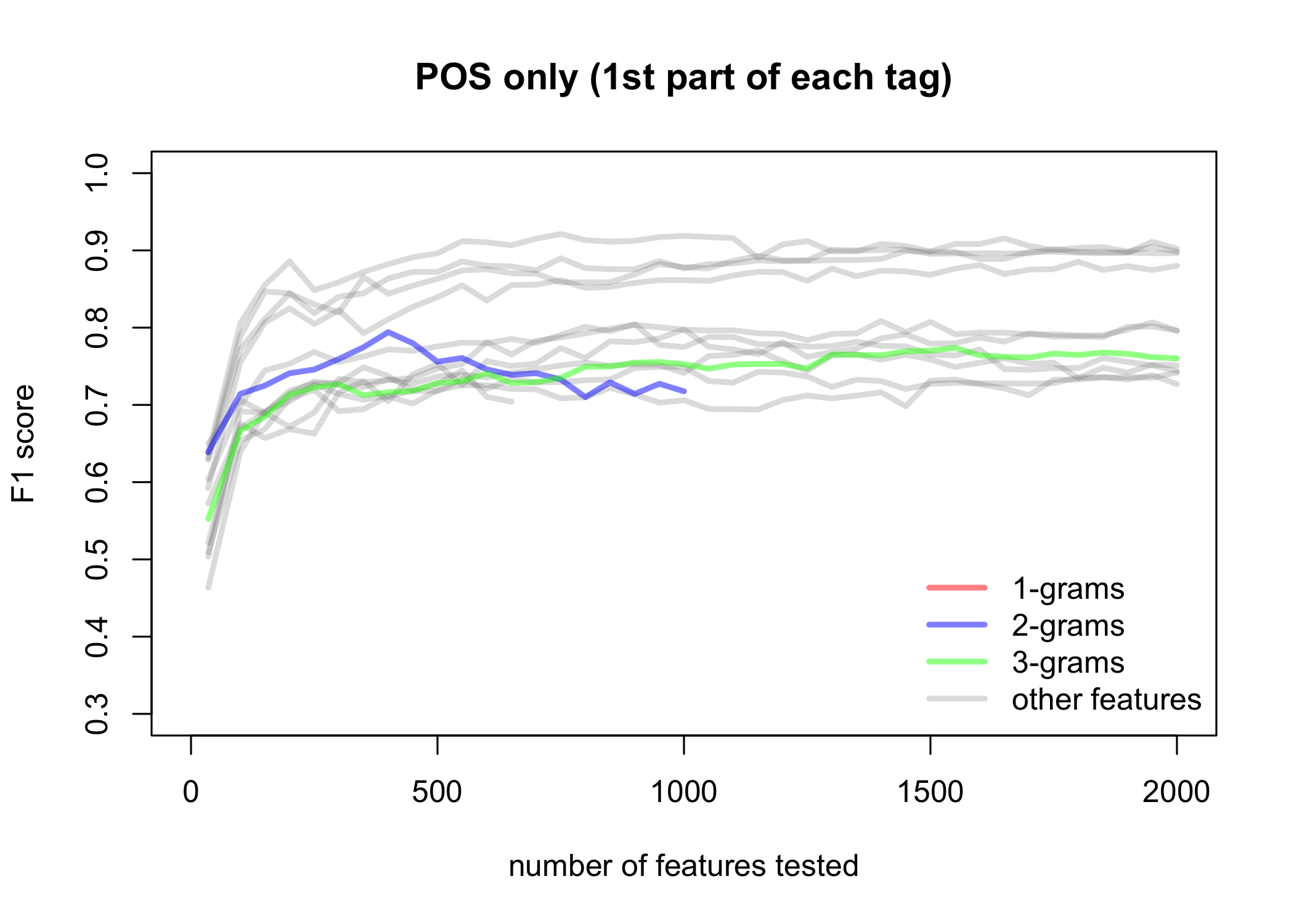}
\caption{Performance of the initial segment of each tag
(i.e.~POS in the strict sense) in the corpus of 189 novels, assessed by
Cosine Delta.}
\end{figure}

The relatively good performance of higher-order POS-tag \emph{n}-grams
over single items or 2-grams deserves a linguistic interpretation. It
clearly shows that syntax (if we believe that it is reflected by
sequences of 3 subsequent POS labels) plays a considerable role in the
authorial fingerprint, even if it cannot compete with the overwhelming
performance of frequent words. Being less noticeable, however, syntactic
style-markers are very stable in terms of resistance to the number of
analyzed \emph{n}-grams. The F1 attributive score of \emph{ca}. 0.75 for
the worst-case scenario provides us with strong evidence that the
syntactic features retain a considerable amount of the authorial
fingerprint.

\begin{figure}
\centering
\includegraphics[width=1\textwidth]{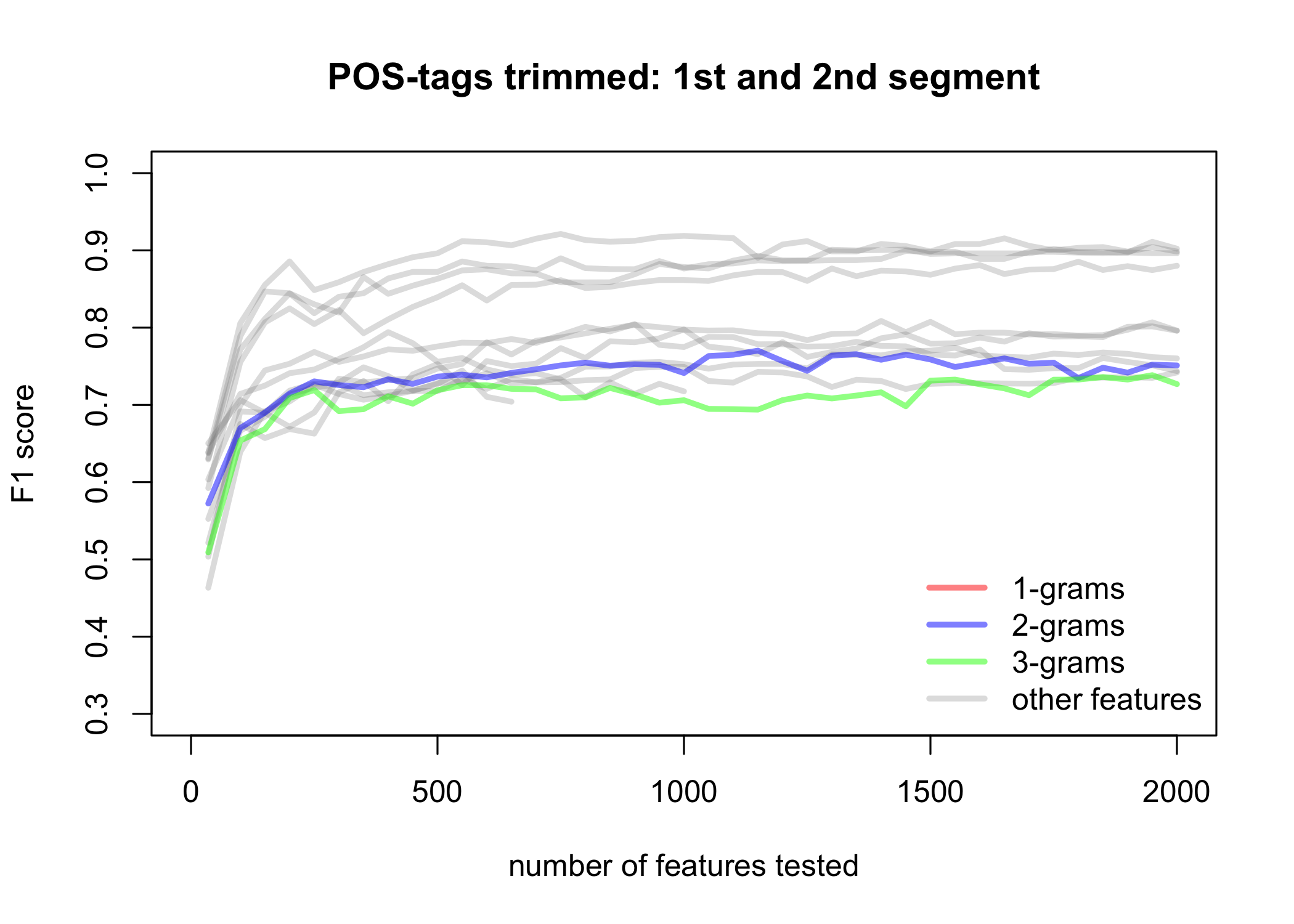}
\caption{Performance of the first initial segments of each tag
in the corpus of 189 novels, assessed by Cosine Delta.}
\end{figure}

\hypertarget{conclusions}{%
\section{Conclusions}\label{conclusions}}

The results obtained in this study allow for a few general observations.
Firstly, this study shows that, at least in Polish, lemmatization is not
necessarily the way to raise attribution accuracy in that language.
Presumably, this claim should be applicable -- by extension -- to other
languages having a rich inflection. This observation is somewhat
counter-intuitive, since lemmatization leads to a decrease of the number
of types and an increase of the number of tokens per type, which in turn
should reduce data sparseness. It turned out otherwise, as if
lemmatization, or a crude way of ``making Polish more like English'',
stripped out some relevant information about authorial uniqueness. Since
we know what exactly is lost in the process of lemmatization, we can
reason that all the suffixes containing inflection play some role in
authorship attribution.

A more convincing evidence of the role of grammar in attribution, is
provided by our tests involving POS-tag \emph{n}-grams. Despite
significantly worse performance, our syntax-based features exhibited a
big potential to distinguish between authors. It is a widely accepted
claim that the linguistic originality of an author manifests itself in
the lexis, i.e.~in predilection to some words and avoidance of other. It
is less obvious whether the same can be said of syntactic constructions;
intuitively, syntax does not allow as much of freedom of choice as
lexis. Our results provide evidence that syntax alone is responsible for
a considerable amount of authorial uniqueness. Even if syntactic
features cannot compete with the lexis, they can still be used as
efficient style-markers, possibly in combination with traditional
features. Interestingly, the loss of accuracy when only grammatical tags
were taken into account was not very high (\emph{ca}. 15\%). This is a
good hint that writers/authors are only slightly less restricted by
syntax than they are by lexis.

\hypertarget{acknowledgements}{%
\section{Acknowledgements}\label{acknowledgements}}

This research was conducted as a result of the project ``Large-Scale
Text Analysis and Methodological Foundations of Computational
Stylistics'' (2017/26/E/HS2/01019) supported by Poland's National
Science Centre.

\hypertarget{references}{%
\section*{References}\label{references}}
\addcontentsline{toc}{section}{References}

\hypertarget{refs}{}
\begin{CSLReferences}{1}{0}
\leavevmode\hypertarget{ref-acedanskiMorphosyntacticBrillTagger2010}{}%
\textbf{Acedański, S.} (2010). A morphosyntactic {Brill} tagger for
inflectional languages. \emph{Advances in {Natural Language
Processing}}. {Reykjavik}, pp. 3--14.

\leavevmode\hypertarget{ref-baayenOutsideCaveShadows1996}{}%
\textbf{Baayen, H., Van Halteren, H. and Tweedie, F.} (1996). Outside
the cave of shadows: Using syntactic annotation to enhance authorship
attribution. \emph{Literary and Linguistic Computing}, \textbf{11}(3):
121--32.

\leavevmode\hypertarget{ref-burrowsTiptoeingInfiniteTesting1996}{}%
\textbf{Burrows, J.} (1996). Tiptoeing into the infinite: {Testing} for
evidence of national differences in the language of {English} narrative.
In Hockey, S. and Ide, N. (eds), \emph{Research in {Humanities
Computing} 4}. {Oxford}: {Oxford University Press}, pp. 1--33.

\leavevmode\hypertarget{ref-burrowsDeltaMeasureStylistic2002}{}%
\textbf{Burrows, J.} (2002). {``{Delta}''}: A measure of stylistic
difference and a guide to likely authorship. \emph{Literary and
Linguistic Computing}, \textbf{17}(3): 267--87.

\leavevmode\hypertarget{ref-ederStylemarkersAuthorshipAttribution2011}{}%
\textbf{Eder, M.} (2011). Style-markers in authorship attribution: A
cross-language study of the authorial fingerprint. \emph{Studies in
Polish Linguistics}, \textbf{6}: 99--114.

\leavevmode\hypertarget{ref-ederMindYourCorpus2013}{}%
\textbf{Eder, M.} (2013). Mind your corpus: Systematic errors in
authorship attribution. \emph{Literary and Linguistic Computing},
\textbf{28}(4): 603--14.

\leavevmode\hypertarget{ref-ederStylometryPackageComputational2016}{}%
\textbf{Eder, M., Rybicki, J. and Kestemont, M.} (2016). Stylometry with
{R}: A package for computational text analysis. \emph{R Journal},
\textbf{8}(1): 107--21.

\leavevmode\hypertarget{ref-evertUnderstandingExplainingDelta2017}{}%
\textbf{Evert, S., Proisl, T., Jannidis, F., Reger, I., Pielström, S.,
Schöch, C. and Vitt, T.} (2017). Understanding and explaining {Delta}
measures for authorship attribution. \emph{Digital Scholarship in the
Humanities}, \textbf{32}(suppl. 2): 4--16
doi:\href{https://doi.org/10.1093/llc/fqx023}{10.1093/llc/fqx023}.

\leavevmode\hypertarget{ref-hirstBigramsSyntacticLabels2007}{}%
\textbf{Hirst, G. and Feiguina, O.} (2007). Bigrams of syntactic labels
for authorship discrimination of short texts. \emph{Literary and
Linguistic Computing}, \textbf{22}(4): 405--17.

\leavevmode\hypertarget{ref-jockersComparativeStudyMachine2010}{}%
\textbf{Jockers, M. L. and Witten, D. M.} (2010). A comparative study of
machine learning methods for authorship attribution. \emph{Literary and
Linguistic Computing}, \textbf{25}(2): 215--23.

\leavevmode\hypertarget{ref-juolaCrosslinguisticTransferrenceAuthorship2009}{}%
\textbf{Juola, P.} (2009). Cross-linguistic transferrence of authorship
attribution, or why {English}-only prototypes are acceptable.
\emph{Digital {Humanities} 2009: {Conference Abstracts}}. {College Park,
MD}: {University of Maryland}, pp. 162--63.

\leavevmode\hypertarget{ref-kennyComputationStyleIntroduction1982}{}%
\textbf{Kenny, A.} (1982). \emph{The Computation of Style: An
Introduction to Statistics for Students of Literature and Humanities}.
{Oxford; New York}: {Pergamon Press}.

\leavevmode\hypertarget{ref-kestemontFunctionWordsAuthorship2014}{}%
\textbf{Kestemont, M.} (2014). Function words in authorship attribution:
From black magic to theory? \emph{Proceedings of the 3rd {Workshop} on
{Computational Linguistics} for {Literature} ({CLFL})}. {Gothenburg,
Sweden}: {Association for Computational Linguistics}, pp. 59--66.

\leavevmode\hypertarget{ref-koppelComputationalMethodsAuthorship2009}{}%
\textbf{Koppel, M., Schler, J. and Argamon, S.} (2009). Computational
methods in authorship attribution. \emph{Journal of the American Society
for Information Science and Technology}, \textbf{60}(1): 9--26.

\leavevmode\hypertarget{ref-luyckxEffectAuthorSet2011}{}%
\textbf{Luyckx, K. and Daelemans, W.} (2011). The effect of author set
size and data size in authorship attribution. \emph{Literary and
Linguistic Computing}, \textbf{26}(1): 35--55.

\leavevmode\hypertarget{ref-mckennaBeckettTrilogyComputational1999}{}%
\textbf{McKenna, W., Burrows, J. and Antonia, A.} (1999). Beckett's
trilogy: Computational stylistics and the nature of translation.
\emph{Revue Informatique Et Statistique Dans Le Sciences Humaines},
\textbf{35}: 151--71.

\leavevmode\hypertarget{ref-przepiorkowskiNarodowyKorpusJezyka2012}{}%
\textbf{Przepiórkowski, A., Bańko, M., Górski, R. L. and
Lewandowska-Tomaszczyk, B. (eds).} (2012). \emph{Narodowy {Korpus Języka
Polskiego}}. {Warszawa}: {PWN}.

\leavevmode\hypertarget{ref-rybickiSuccessRatesMostfrequentwordbased2015}{}%
\textbf{Rybicki, J.} (2015a). Success {Rates} in
most-frequent-word-based authorship attribution: {A} case study of 1000
{Polish} novels from {Ignacy Krasicki} to {Jerzy Pilch}. \emph{Studies
in Polish Linguistics}, \textbf{10}(2): 87--104.

\leavevmode\hypertarget{ref-rybickiViveDifferenceTracing2015}{}%
\textbf{Rybicki, J.} (2015b). Vive la différence: {Tracing} the
(authorial) gender signal by multivariate analysis of word frequencies.
\emph{Digital Scholarship in the Humanities}, \textbf{31}(4): 746--61
doi:\href{https://doi.org/10.1093/llc/fqv023}{10.1093/llc/fqv023}.

\leavevmode\hypertarget{ref-rybickiDeeperDeltaGenres2011}{}%
\textbf{Rybicki, J. and Eder, M.} (2011). Deeper {Delta} across genres
and languages: Do we really need the most frequent words? \emph{Literary
and Linguistic Computing}, \textbf{26}(3): 315--21.

\leavevmode\hypertarget{ref-sokolovaSystematicAnalysisPerformance2009}{}%
\textbf{Sokolova, M. and Lapalme, G.} (2009). A systematic analysis of
performance measures for classification tasks. \emph{Information
Processing and Management}, \textbf{45}(4): 427--37
doi:\href{https://doi.org/10.1016/j.ipm.2009.03.002}{10.1016/j.ipm.2009.03.002}.

\leavevmode\hypertarget{ref-stamatatosSurveyModernAuthorship2009}{}%
\textbf{Stamatatos, E.} (2009). A survey of modern authorship
attribution methods. \emph{Journal of the American Society for
Information Science and Technology}, \textbf{60}(3): 538--56.

\leavevmode\hypertarget{ref-wiersmaAutomaticallyExtractingTypical2011}{}%
\textbf{Wiersma, W., Nerbonne, J. and Lauttamus, T.} (2011).
Automatically extracting typical syntactic differences from corpora.
\emph{Literary and Linguistic Computing}, \textbf{26}(1): 107--24.

\end{CSLReferences}

\bibliographystyle{unsrt}
\bibliography{bibliography.bib}

\end{document}